\documentclass[a4paper, 10pt, conference]{ieeeconf}      % Use this line for a4 paper todo 

\makeatletter
\def\endthebibliography{%
	\def\@noitemerr{\@latex@warning{Empty `thebibliography' environment}}%
	\endlist
}
\makeatother %% a workaround from https://tex.stackexchange.com/questions/350907/error-latex-error-somethings-wrong-perhaps-a-missing-item

\IEEEoverridecommandlockouts                              % This command is only needed if 
\usepackage{cite}
\usepackage{amsmath,amssymb,amsfonts}
\usepackage{algorithmic}
\usepackage{graphicx}
\usepackage{textcomp}
\usepackage{xcolor}
\usepackage{pgfplots}
\usepackage{url} % new package
\usepackage{makecell} % new pacakge
\usepackage{caption}
\usepackage{subcaption}
\usepackage[flushleft]{threeparttable}
\usepackage{tabularx}
\usepackage{booktabs}
\usepackage{array}

\newcolumntype{P}[1]{>{\centering\arraybackslash}p{#1}}

\title{\LARGE \bf
	Overview of Tools Supporting Planning for Automated Driving 
}

\author{Kailin Tong$^{1}$, Zlatan Ajanovic$^{1}$ and Georg Stettinger$^{1}$  % <-this % stops a space
	\thanks{$^{1}$ K. Tong, Z. Ajanovic and G. Stettinger are with Virtual Vehicle Research Center, Inffeldgasse 21a, 8010 Graz, Austria. {\tt\small firstname.familyname@v2c2.at}
	}
}
\begin{document}

	\maketitle
	\thispagestyle{empty}
	\pagestyle{empty}

	%%%%%%%%%%%%%%%%%%%%%%%%%%%%%%%%%%%%%%%%%%%%%%%%%%%%%%%%%%%%%%%%%%%%%%%%%%%%%%%%
	\begin{abstract}
	Planning is an essential topic in the realm of automated driving. Besides planning algorithms that are widely covered in the literature, planning requires different software tools for its development, validation, and execution. This paper presents a survey of such tools including map representations, communication, traffic rules, open-source planning stacks and middleware, simulation, and visualization tools as well as benchmarks. We start by defining the planning task and different supporting tools. Next, we provide a comprehensive review of state-of-the-art developments and analysis of relations among them. Finally, we discuss the current gaps and suggest future research directions.
		
	\end{abstract}

	%%%%%%%%%%%%%%%%%%%%%%%%%%%%%%%%%%%%%%%%%%%%%%%%%%%%%%%%%%%%%%%%%%%%%%%%%%%%%%%%
	\section{INTRODUCTION} \label{INTRODUCTION} % update

	As an emerging technology, automated driving has been the object of great research efforts in the last several decades. These developments have been fueled by introduction of advanced sensing technology and high-performance computation hardware together with the potential disruptive impact on transportation systems and social benefits (e.g. 94 \% of motor vehicle crashes are caused by human mistakes \cite{NationalHighwayTrafficSafetyAdministrationandU.S.DepartmentofTransportation.}.) Autonomous vehicles (AVs) have the potential to dramatically enhance road safety, provide efficient traffic flow and reduce emissions, while improving mobility and  general well-being \cite{Crayton.2017}, \cite{safety_first}. However, many obstacles to mass-market penetration of AVs remain, such as technical liability, cost, licensing, security and privacy concerns \cite{Fagnant.2015}.
	
	Planning is a critical part to realize driving autonomy, incorporated with perception and execution within the system architecture \cite{Watzenig.2017}. Existing planning algorithms originate predominantly from the community of robotics: \textit{Their target is to convert high-level specifications of tasks from humans into low-level descriptions of how to move}  \cite{lavalle2006planning}. The major task of planning for automated driving is to generate a collision-free and feasible path or trajectory towards a destination that considers vehicle dynamics, maneuver abilities, traffic rules, road boundaries or any other constraints, while optimizing driving targets \cite{Katrakazas.2015}.  
	
	Many publications provide an overview of planning approaches for automated driving with respect to a system-to-component structure: Pendleton et al. discussed developments in the realm of autonomous vehicle software systems and their components \cite{Pendleton.2017}. Badue et al. surveyed about self-driving cars (SAE level 3 or higher) \cite{badue2019self}. Yurtsever et al. gave an overall survey of automated driving systems and highlighted the emerging technologies \cite{Yurtsever.12.06.2019}. Control and planning architecture for connected and automated vehicles are reviewed by Guanettia et al., in which it states that cloud service can remotely perform a part of planning computation, e.g. (dynamic) routing and long-term trajectory optimization \cite{Guanetti.11.04.2018}. For review about functional system architectures of automated driving, the interested reader is referred to \cite{Tas.19.06.201622.06.2016},\cite{Matthaei.2015},\cite{Ulbrich.24.03.2017}. Many works survey the planning algorithms, which are summarized in Section \ref{DEFINITIONS}.
	
	Along with the booming of planning-related research, a large number of supportive software or hardware arise. For the DARPA urban challenge in 2007, different teams proposed different software infrastructures. The winner of DARPA Urban Challenge, Tartan Racing team attributed their accomplishment partially to the supporting tools: \textit{The study of mobile robot software infrastructure is important because a well-crafted infrastructure has the potential to speed the entire robot development project by enabling the higher level architecture} \cite{mcnaughton2008software}. Other noteworthy teams also developed their own simulation systems, e.g., Talos from MIT, Junior from Stanford University and Odin from Virginia Tech \cite{Buehler.2009}. 
	
	Different surveys focus on some aspects of supporting tools. Particularly, the interested reader is referred to the literature about different models of maps\cite{Li.06.03.201708.03.2017}, high definition maps\cite{Liu.2019}, simulators \cite{Yurtsever.12.06.2019},\cite{JeffCraighead.2007},\cite{pilz2019development} or specifically traffic simulation \cite{Chao.2019}, and datasets \cite{Yurtsever.12.06.2019}. Explicit definitions of scene, situation and scenario for testing of planning modules can be found in \cite{Ulbrich.15.09.201518.09.2015}.
	
	This paper pays special attention to the tools supporting planning in a broad sense, which is the first work that comprehensively presents and compares state-of-the-art tools related to planning, and shows the explicit connections and bridges between them. To our knowledge, no other work surveys about this significant problem. The aim of this paper is to fill this gap in the literature with a thorough survey. We believe that our work will facilitate readers who want to quickly establish a platform to test developed planning algorithms.
	
	The remainder of the paper is structured as follows: foundational definitions form the body of Section \ref{DEFINITIONS}; while Section \ref{SUPPORTING TOOLS} presents an introduction of tools applied to planning, followed by their specific characteristics. The possible coupling between different tools or systems is then described in Section \ref{Coupling of Tools}. Finally, in Section \ref{CONCLUSION} the paper discusses remaining challenges and future research directions.
	
	\section{DEFINITIONS} \label{DEFINITIONS}

	\subsection{SENSE-PLAN-ACT} % \cite{murphy2019introduction}
	Paradigms in robotics follow three commonly accepted primitives: SENSE, PlAN and ACT \cite{russell2016artificial} (see the center of Figure \ref{fig:tools}). The functions of an autonomous vehicle can be divided into these three general categories as well. If a function is acquiring information about the environment using vehicle's sensors and producing a world model for other functions, then that function falls in the SENSE category. If a function is receiving the aforementioned world model and producing one or more action plans for the vehicle to perform, that function is in the PLAN category. Functions, which generate actuator (e.g. steering and E-motor) commands according to the directives derived from the planning stage, fall into ACT \cite{Silva}. 
	
	\subsection{Typical Planning System} \label{Typical Planning System}
	A typical planning system for autonomous vehicles is hierarchically decomposed into four classes, as proposed by Varaiya in 1993 \cite{Varaiya1993195}: (1) route planning layer, (2) path planning layer, (3) maneuver choice layer, (4) trajectory planning layer (originally called control planning in \cite{Varaiya1993195}, and termed as motion planing in \cite{Paden.2016}). Planning on the highest level for a route through the road network has been heavily studied. For a most recent comparison of practical routing algorithms that can be applied for both conventional and self-driving vehicles, see \cite{Bast.20.04.2015}. Path/trajectory planning as well as behavior planning are normally incorporated in literature reviews. After the famed DARPA urban challenge in 2007\cite{Buehler.2009}, different surveys related to this topic argue that their novelty lies in various concentration, including:  the pioneer considering planning in all three levels (i.e. path, behavior, trajectory)\cite{Katrakazas.2015}, presenting the state of the art about planning strategies \cite{Gonzalez.2016}, survey of lane change and merge maneuvers for CAVs \cite{Bevly.2016}, attention to planning and control algorithms regarding the urban setting\cite{Paden.2016}, special interest of methods for sampling-based planning with constraints \cite{kingston2018sampling}, focus on autonomous overtaking\cite{Dixit.2018} and emphasis on integration of perception and planning as well as behavior-aware planning \cite{Schwarting.2018}. Specific survey about prediction models can be seen in \cite{lefevre2014survey}. In addition, Ilievski et al. explored design space of behavior planning for autonomous vehicles and stated the future research direction: learned systems supervised by programmed logic\cite{Ilievski.21.08.2019}.
			
%	\begin{figure}[thpb]
%		\centering
%		\includegraphics[scale=0.5]{pictures/SensePlanAct.png}
%		\caption{Hierarchical paradigm with SENSE-PLAN-ACT cycle\cite{murphy2019introduction}}
%		\label{SensePlanAct}
%	\end{figure}

	\subsection{Tools Supporting Planning} % copy here	
	In this paper we spotlight automated driving functions in the PLAN class, e.g. route planing, path or trajectory planning and behavior planing probably in cooperation with a prediction module. Although the existing end-to-end approach \cite{bojarski2016end,rausch2017learning,jaritz2018end}, which integrates SENSE, PLAN and even ACT, is a vast area of research and has claimed potential to improve robustness, we also leave the kind of approach and its supporting tools unanalyzed. These following components are essential for implementation and testing of a classical planing algorithm: An environment integrating traffic rules, the data structure representing the environment and the mission, and benchmarks (probably incorporating datasets) that generate a specific environment and evaluate the performance of planning algorithm. Some middleware or communication modules serve as interfaces between different data structures, programming languages or software components, and hence we attach importance to them. Besides, many open-source planning stacks are also reviewed in survey as they can support researchers to fast iterate their devolved algorithms. According to the proposed functional system architecture in \cite{Ulbrich.24.03.2017}, tools supporting planning hence comprise of: maps, communication, traffic rules, middleware, simulators and benchmarks.
	
	\begin{figure}[t]
	\centering
	\includegraphics[scale=0.45]{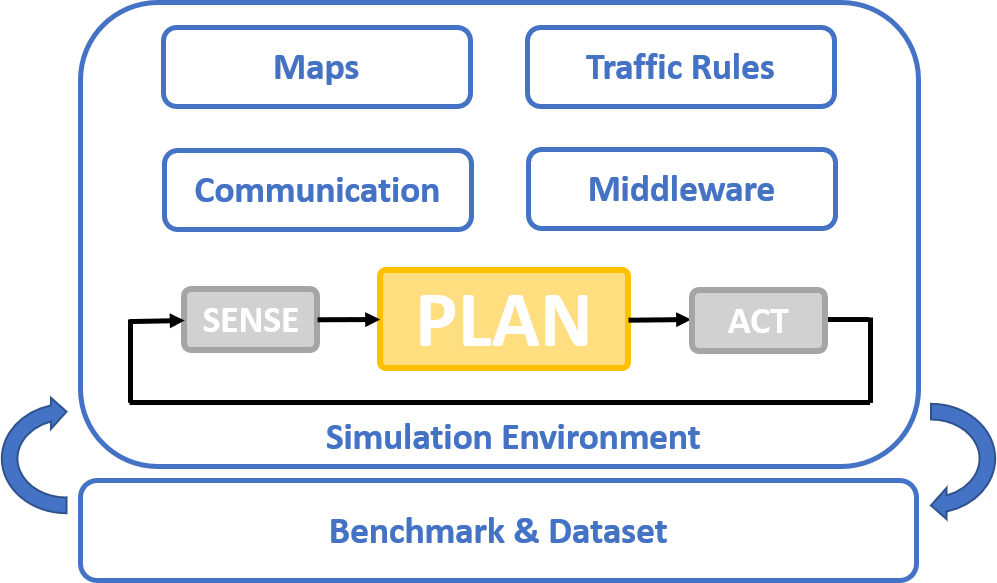}
	\caption{Tools supporting planning}
	\label{fig:tools}
	\end{figure}	
	
	\section{SUPPORTING TOOLS} \label{SUPPORTING TOOLS}
	 The tools that are covered in this paper are shown in Figure \ref{fig:tools}. In addition, Data visualization is powerful for debugging and validating automated driving functions. A few famous data visualization platforms are reviewed as well.  
	
	\subsection{Maps}  % Note update here	
\begin{figure*}
	\centering
	\begin{subfigure}[b]{0.18\textwidth}
		\centering
		\includegraphics[width=\linewidth]{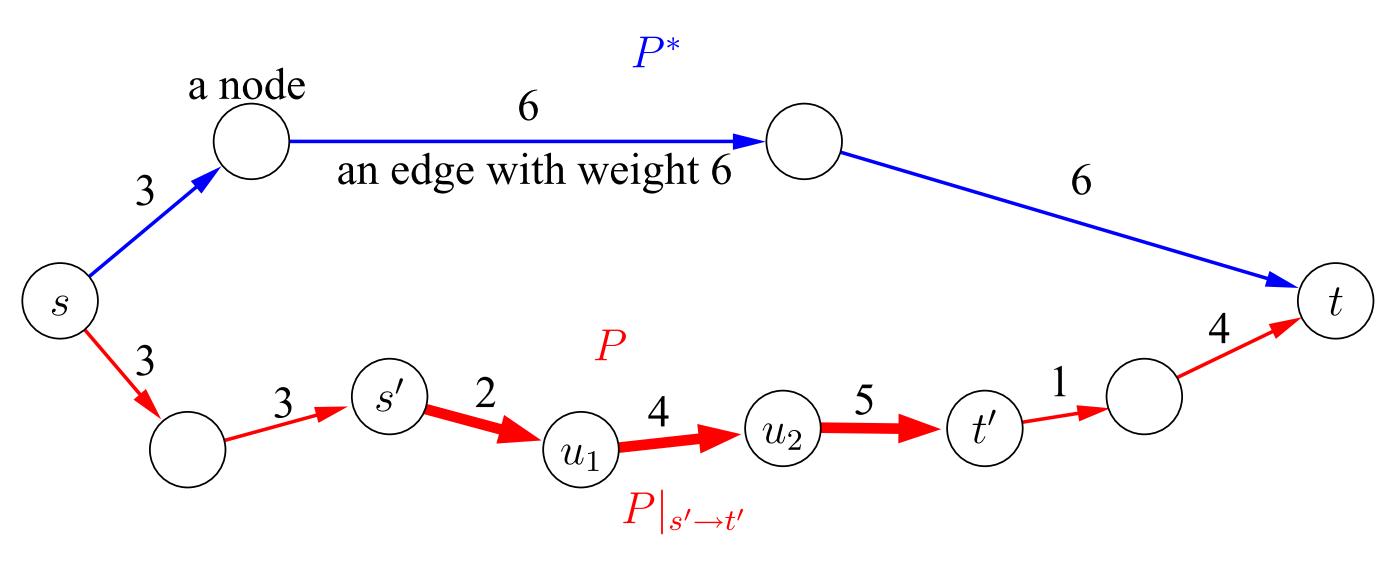}
		\caption{Directed Graph \cite{Schultes}}
		\label{fig:directed_graph}
	\end{subfigure}\hfill
	\begin{subfigure}[b]{0.18\textwidth}
		\centering
		\includegraphics[width=\linewidth]{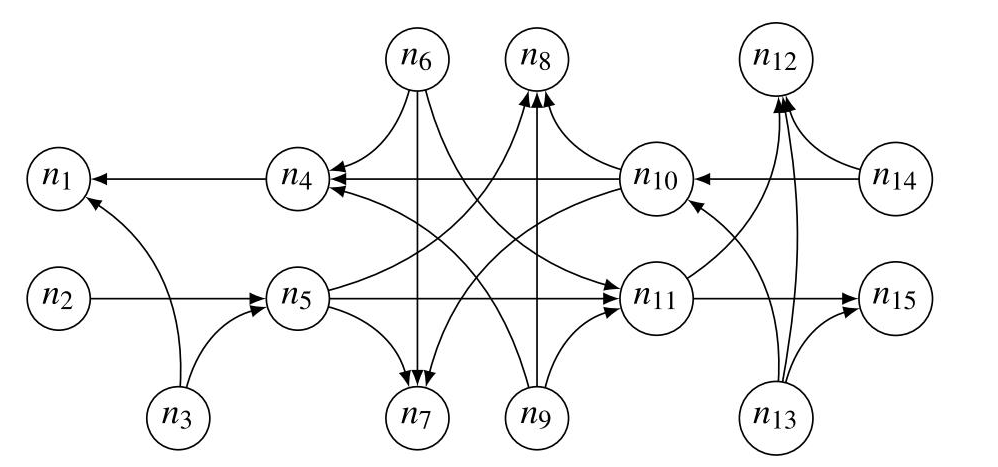}
		\caption{Adjoint Graph \cite{Eskandarian.2012}}
		\label{fig:adjoint_graph}
	\end{subfigure}\hfill
	\begin{subfigure}[b]{0.18\textwidth}
		\centering
		\includegraphics[width=\linewidth]{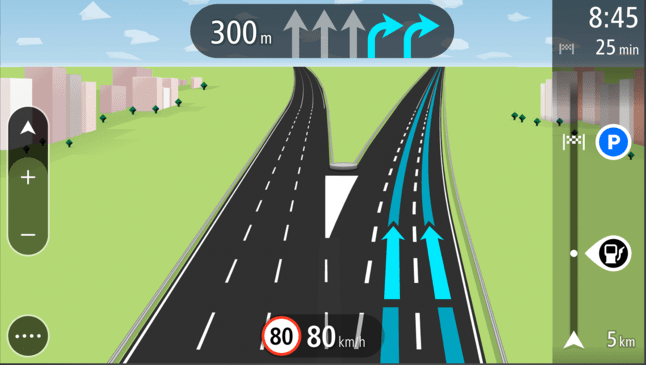}
		\caption{Center Lines \cite{tomtom}}
		\label{fig:tomtom2}
	\end{subfigure}\hfill
	\begin{subfigure}[b]{0.18\textwidth}
		\centering
		\includegraphics[width=\linewidth]{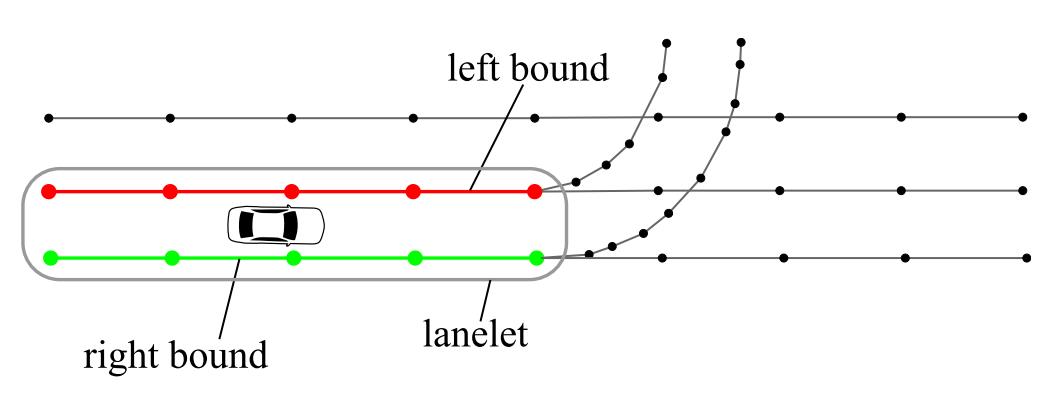}
		\caption{Bounds \cite{Lanelets}}
		\label{fig:lanelets}
	\end{subfigure}
	\begin{subfigure}[b]{0.18\textwidth}
	\centering
	\includegraphics[width=\linewidth]{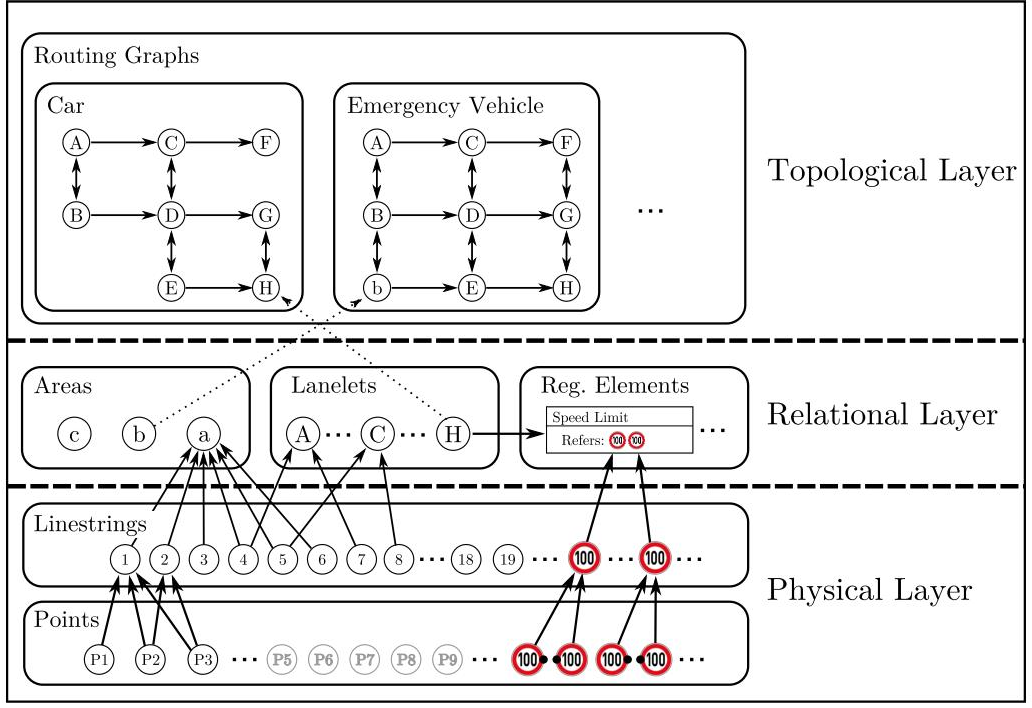}
	\caption{Multiple Layers \cite{Lanelet2}}
	\label{fig:lanelet2}
	\end{subfigure}
	\begin{subfigure}[b]{0.16\textwidth}
	\centering
	\includegraphics[width=\linewidth]{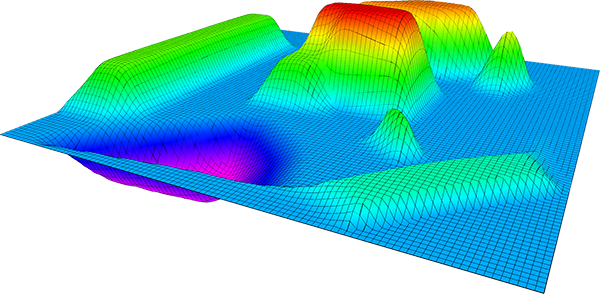}
	\caption{Grid Map \cite{Fankhauser2016GridMapLibrary}}
	\label{fig:grid_map_rviz_plugin_example}
	\end{subfigure}
	\begin{subfigure}[b]{0.16\textwidth}
	\centering
	\includegraphics[width=\linewidth]{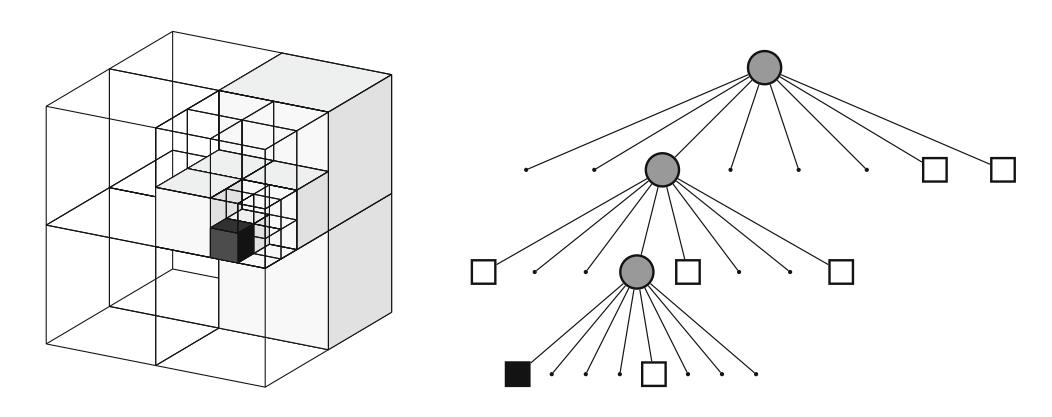}
	\caption{Octotree \cite{hornung2013octomap}}
	\label{fig:octotree}
	\end{subfigure}
	\begin{subfigure}[b]{0.165\textwidth}
	\centering
	\includegraphics[width=\linewidth]{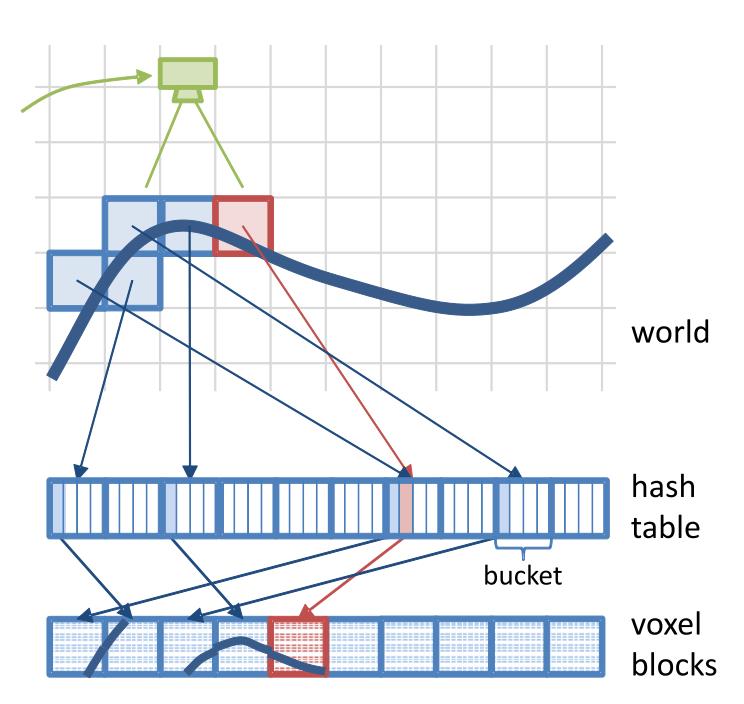}	
	\caption{Voxel Hashing \cite{niessner_2017}}
	\label{fig:voxel_hashing}
	\end{subfigure}
	\begin{subfigure}[b]{0.16\textwidth}
	\centering
	\includegraphics[width=\linewidth]{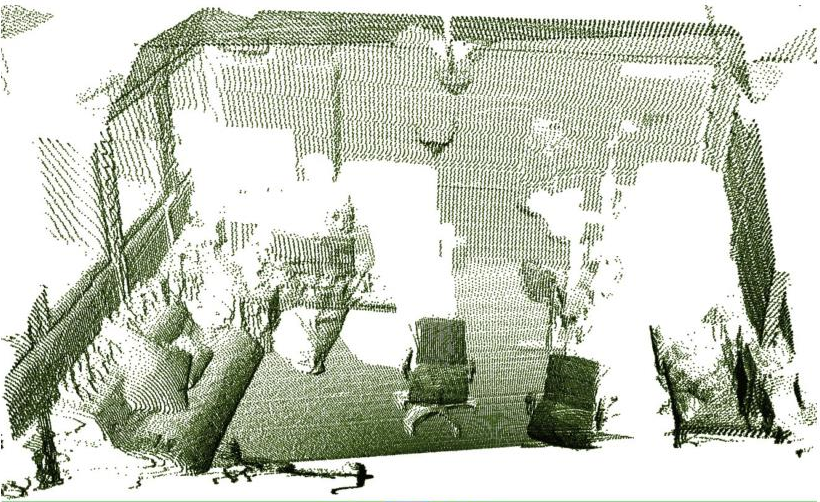}	
	\caption{Point Cloud \cite{Rusu_ICRA2011_PCL}}
	\label{fig:pcl}
	\end{subfigure}
	\begin{subfigure}[b]{0.16\textwidth}
	\centering
	\includegraphics[width=\linewidth]{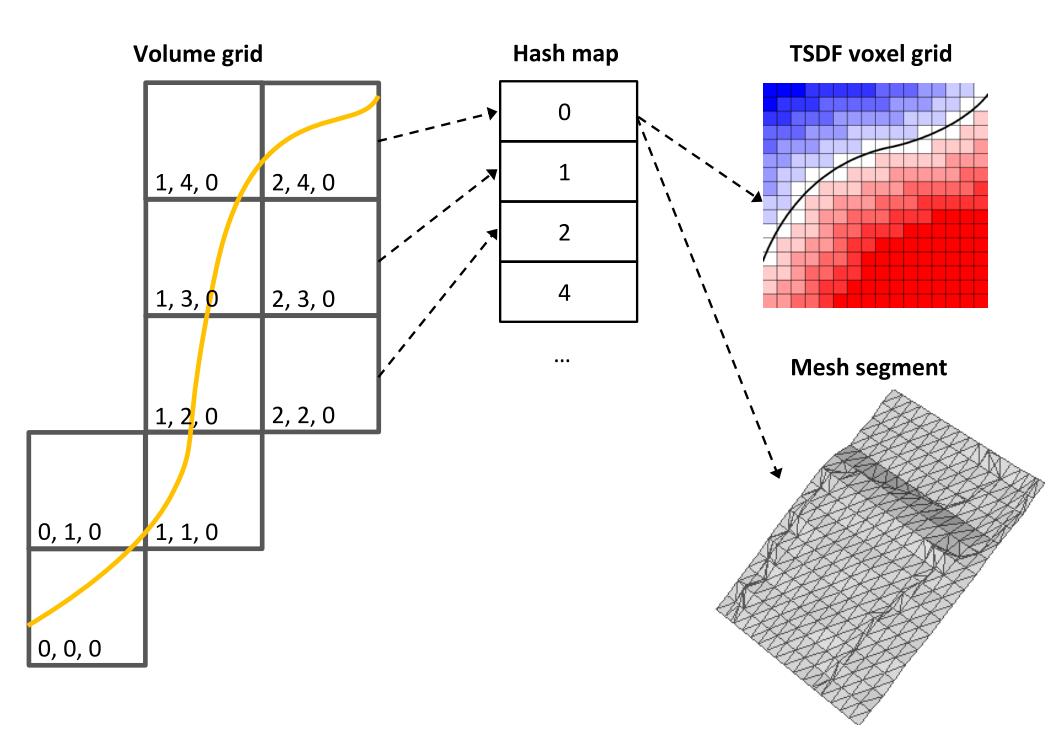}
	\caption{TSDF \cite{Klingensmith2015ChiselRT}}
	\label{fig:TSDF}
	\end{subfigure}
	\begin{subfigure}[b]{0.16\textwidth}
	\centering
	\includegraphics[width=\linewidth]{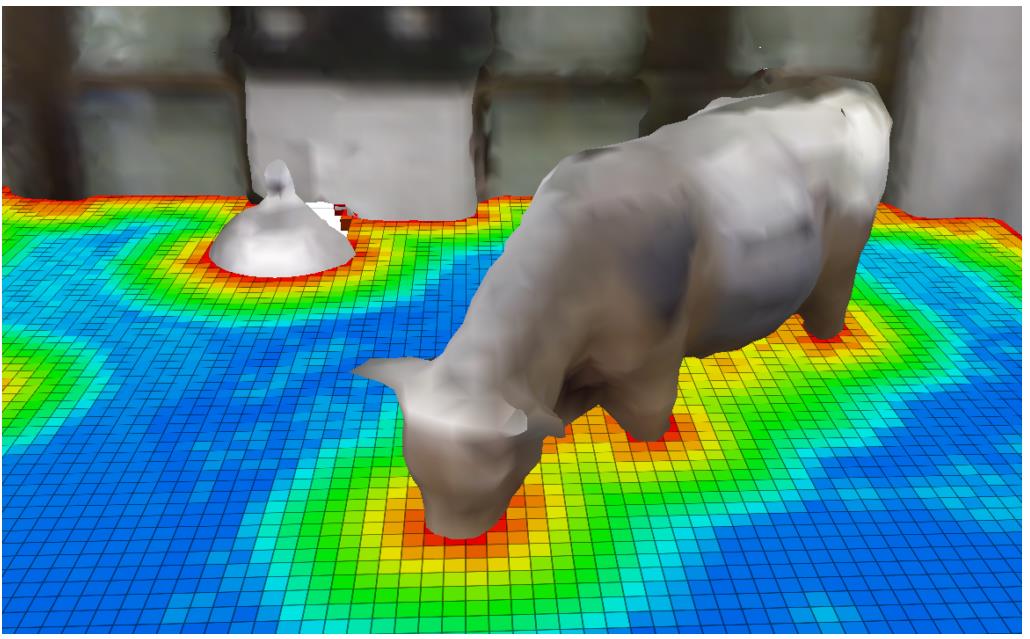}
	\caption{ESDF \cite{Oleynikova_2017}}
	\label{fig:TSDF}
	\end{subfigure}
	\caption{Map representations}
	\label{fig:map}
\end{figure*}

	The earliest in-car navigation map was reported to be rolled paper maps in an Iter Avto in 1930 \cite{newcomb_2018}. As more details are incorporated in maps, the map type has evolved from paper map to digital map, enhanced digital map and recent High Definition (HD) map. The planning tasks with different targets entail map models with different level of details. HD map provides the most sufficient information and can be generally categorized into three layers \cite{Liu.2019}: road model, lane model and localization model. The road model is mainly used for strategic planning (navigation) and supports object prediction and behavior generation. The lane model is used for tactical planning (guidance) and motion planning. The localization model, which provides direct access to elements, is most relevant to localization \cite{Li.06.03.201708.03.2017} but it can also strengthen planning in an unstructured and complex environment. Examples using this model can be found in robotics, such as occupancy grid map, voxel hashing, octotree, point clound map, Truncated Signed Distance Functions (TSDF) map and Euclidean Signed Distance Functions (ESDF) map. Different map representations and application examples are shown in Table \ref{tab:map}. An example of each map representation is shown in Figure \ref{fig:map}.

	\begin{table}[htp]
		\caption{Map representations for planning research}
		\begin{center}
			\begin{tabular}{ | c | c | c | } %% todo: open-source
				\hline
				%			Simulator & Traffic Flow  & Interactive  & Perception & Vehicle \\ 
				%					 &  & Road User  & input & Dynamics \\ \hline \hline 		
				\makecell[c]{Map Layer} & \makecell[c]{Representation} & \makecell[c]{Examples}   \\ \hline \hline
				Road   & directed graph & a graph for routing  \cite{Schultes}  \\ \hline 
				Road   & adjoint graph & \makecell{a graph for routing\\and cost prediction \cite{Nunzio2017}}    \\ \hline 						
				Lane   & \makecell{center lines\\ with attributes} & \makecell{OSM \cite{osm}, OpenDRIVE \cite{opendrive}\\TomTom \cite{tomtom}, Here \cite{here_2019}}  \\ \hline 
				Lane & \makecell{geometric bounds \\ by polygonal lines}  & Lanelets \cite{Lanelets}  \\ \hline 	
				\makecell{Road+Lane+\\Localization} & \makecell{multiple-layer \\ structure} & \makecell{Lanelet2 \cite{Lanelet2}, Here \cite{here_2019} \\ Lyft \cite{kumar_chellapilla_2018}, TomTom \cite{tomtom} } \\ \hline	
				Localization   & \makecell{occupancy\\grid map}  & Grid Map \cite{Fankhauser2016GridMapLibrary}  \\ \hline 	
				Localization   & voxel hashing &  \makecell{voxel hashing \cite{niessner_2017}\\ InfiniTAM \cite{InfiniTAM_ISMAR_2015}}   \\ \hline 						
				Localization   & \makecell{octotree}  & OctoMap \cite{hornung2013octomap}  \\ \hline 
				Localization   & \makecell{point cloud map}  & PCL \cite{Rusu_ICRA2011_PCL} \\ \hline 	
				Localization   & \makecell{TSDF map}  & OpenChisel \cite{Klingensmith2015ChiselRT}  \\ \hline 	
				Localization   & \makecell{ESDF map}  & \makecell{VoxBlox \cite{Oleynikova_2017}, FIESTA \cite{Han_ESDF}\\TRR's Local Map \cite{gao2019optimal}}  \\				

				 \hline 																															
			\end{tabular}
			\label{tab:map}
		\end{center}
	\end{table}
	
	\subsection{Communication}
	The planning stack might have communication interfaces with other traffic participants as well as infrastructure via Vehicle-to-everything  (V2X) communication, or with operators or passengers via human-machine interface (HMI) \cite{Ulbrich.24.03.2017}. The HMI or V2X gives navigation, mission or guidance inputs to the different level of planing algorithms. Hence simulation of the communication is essential in order to realize the full planning cycle. Some existing network simulators can be coupled with the other domain's simulation tools, e.g. OMnet++ simulator \cite{Varga} can be coupled with SUMO \cite{lopez2018microscopic} via Veins \cite{Sommer}. However, due to the limited pages, the sector of HMI as well as V2X research will not be further discussed in this work. % For readers who are interested in technical details of V2X communication, see \cite{ALSULTAN2014380} \cite{Lu.2014}, for those who want to go deep into HMI design, see \cite{hmi}.
	
	\subsection{Traffic Rules}
	
	Autonomous vehicles generally have to operate in an environment populated with other traffic participants (generally human drivers). Human interactions in traffic are governed by traffic rules. It is challenging to express traffic rules in a form understandable by an algorithm. A little research has been done on formalizing traffic rules. One of the first efforts in this direction was presented in \cite{rizaldi2015formalising}. There, traffic rules are formalized in higher order logic using the Isabelle theorem prover. In this way, it is possible to check the compliance of traffic rules unambiguously and formally for vehicle trajectory validation. The other approach is to represent traffic rules geometrically as obstacles in a configuration space of motion planning problem. Such approach was introduced in \cite{ajanovic2018search} and further extended in \cite{ajanovic2019towards}. There different traffic rules, such as speed limits, prohibited lane change, traffic lights, etc., are represented geometrically in 3D space.
	In some occasions, it is necessary to violate some rules during driving for achieving higher goals (i.e. avoiding collision). To deal with this problem, authors in \cite{censi2019liability} introduced a rule book with hierarchical arrangement of different rules. Finally, it is important to note the connection between traffic rules and maps: the map (i.e. Lanelet 2 \cite{Lanelet2}) should provide information about locally applicable rules, traffic signs etc.
	
	\subsection{Open-Source Planning Stacks} \label{ssec:planningStacks}
	
	Many researchers have contributed open-source planning software to the automated driving community. ROS (Robot Operating System) incorporates many motion planning or navigation packages, such as Open Motion Planning Library (OMPL) \cite{sucan2012open}, MoveIt \cite{chitta2012moveit}, navigation package \cite{martinez2013learning} and teb local planner \cite{rosmann2017integrated}, mainly applied in robotics. Another collections of robotics algorithms including those for planning are PythonRobotics \cite{sakai2018pythonrobotics} and CppRobotics \cite{onlytailei_2019}.
	The OpenPlanner, which is composed of a global path planner, a behavior state generator and a local planner, has been implemented in the popular open-source software Autoware \cite{autoware}. One competitor to Autoware in the open-source field is the Baidu Apollo autonomous driving platform, in which a real-time motion planning system is integrated \cite{fan2018baidu}. Nvidia DriveWorks \cite{NVIDIA_DRIVE} and openpilot \cite{comma_ai_2019} are another two commonly used software-stacks capable of realizing automated driving on road. 
	
	\subsection{Simulators} % copy here
	A simulator should provide manifold information for the planning stack: road network, traffic data for route planning, sufficient data for perception to generate a \textit{Scene} (interface between perception and behavior, defined in \cite{Ulbrich.15.09.201518.09.2015}) for behavior planning, and environment features for trajectory planning \cite{Ulbrich.24.03.2017}. As an automated driving system is a highly complex and coupled system, sensor data for SENSE and ACT are also required. In Table \ref{tab:simVsSuitability}, the simulation tools commonly used are listed.  The second column of the table shows the focused realm of each simulator, although the trend is that one simulator supports multi-domain simulations. VD is interpreted again as Vehicle Dynamics, while AD is the abbreviated form of automated driving.
			
	\begin{table}[htp]
		\caption{Simulators for planing, inspired by \cite{pilz2019development}}
		\begin{center}
			\begin{tabular}{  | c | c | c | c | c | c | c | c | } %% todo: open-source
				\hline
				%			Simulator & Traffic Flow  & Interactive  & Perception & Vehicle \\ 
				%					 &  & Road User  & input & Dynamics \\ \hline \hline 		
				\makecell[c]{Simulator} & \makecell[c]{Group} & \makecell[c]{V2X} & \makecell[c]{TF} & \makecell[c]{DM} & \makecell[c]{SE} & \makecell[c]{VI} & \makecell[c]{VD}   \\ \hline \hline
				\textbf{Carla}  \cite{dosovitskiy2017carla} & graphics & i & + & + & + & + & + \\ \hline %% todo: confirm it by API 
				Cognata \cite{cognata} & graphics & o  & + & + & + & ++ & o \\ \hline
%				https://www.youtube.com/watch?v=CDYm-4CAfuU	
				\textbf{LGSVL} \cite{LG_2019} & graphics & i  & + & + & + & + & + \\ \hline		
%				Unity  & graphics & o  & o & o & o & ++ & o \\ \hline	
%				SimViz \cite{} & &   &  &  &  &  &  \\ \hline	lacking info		
				\textbf{Gazebo} \cite{osrf} & robotics & o & o & o & + & o & + \\ \hline
				\textbf{USARSim} \cite{carpin2007usarsim} & robotics & o & o  & o & + & - & o \\ \hline
				\textbf{AirSim} \cite{shah2018airsim} & robotics & o  & o & o & + & + & + \\ \hline % see slide of Yu Huang				
%				PELOPS &   & i & + & + & o &  \\ \hline
				\textbf{MORSE} \cite{morse} & robotics & o  & o & o & + & - & o \\ \hline
				\textbf{TORCS} \cite{Wymann_2015}  & racing & o  & o & + & o & o & +  \\ \hline				
				SynCity \cite{cvedia} & AD &  o & o & o & ++ & ++ & + \\ \hline % todo: lacking information as it is commerialized
				PreScan \cite{tass_international_2018} & AD & ++ & + & + & ++ & - & o \\ \hline % https://tass.plm.automation.siemens.com/prescan-features

				Righthook \cite{righthook} & AD & o & + & ++ & + & + & + \\ \hline % https://righthook.io/solutions/#packages

				SCANeR \cite{avsimulation} & AD & +  & + & + & + & + & + \\ \hline % https://www.avsimulation.fr/applications/
				% todo how good is the SCANeR software? Zlatan
				VTD \cite{vires} & AD & i & + & + & ++ & ++ & + \\ \hline % https://www.youtube.com/watch?v=NksyrGA8Cek				
				\makecell{Autono \\ Vi-Sim \cite{best2018autonovi}}   & AD & o  & + & - & + & + & + \\ \hline
				rFpro \cite{rfpro_2019} & AD & o  & i & i & + & + & ++ \\ \hline
%		http://www.rfpro.com/driving-simulation/traffic-and-pedestrians/					
				Vissim \cite{ptvgroup} & traffic & i  & ++ & + & -{}- & -{}- & -  \\ \hline % Marlies
%				SiVIC &   & i & ++ & ++ & + & ++ \\ \hline
				\textbf{Sumo} \cite{lopez2018microscopic} & traffic & i  & ++ & + & -{}- & -{}- & -{}- \\ \hline
				Aimsun \cite{aimsun} & traffic & i  & ++ & + & -{}- & -{}- & -{}- \\ \hline				
%				RTMaps &   & i & o & -{-} & -{}- &  \\ \hline
%				CVIS &   & i & + & + & -{}- &  \\ \hline
%				RoadView &   & - & -{}- & -{}- & + &  \\ \hline

				CarMaker \cite{ipg} & VD & ++ & + & + & + & + & ++ \\ \hline % https://ipg-automotive.com/areas-of-application/adas-automated-driving/car2x-applications/#smart-home
%			https://ipg-automotive.com/products-services/simulation-software/carmaker/	
							
			\end{tabular}
		    \begin{tablenotes}
			\small
			\item The open-source software's name is in \textbf{bold}. The symbols are rating the quality of implementation: (-{}-) very poor, (-) poor, (o) not rated or irrelevant, (+) good, (++) very good,  (i) some efforts to implement. The abbreviations refer to: TF -- Traffic flow simulation, DM -- Driver model for non-ego objectives, SE -- detail and variety of sensors, VI -- detail of the rendered graphics, VD -- detail of vehicle dynamics.
			\end{tablenotes}
			\label{tab:simVsSuitability}
		\end{center}
	\end{table}
		
	\subsection{Data Visualization} % copy here
	Data Visualization tools are required during the development of algorithms for autonomous vehicles, as they can be used to display how an autonomous vehicle perceives and interprets the world around; it also helps developers to understand the decisions made by the vehicle. Furthermore, it demonstrates how autonomous vehicles work and hence improves users' trust in it. As stated in Section \ref{INTRODUCTION}, various teams in DARPA urban challenge had their unique tools for simulation and visualization. RViz is a popular tool in ROS for visualizing data flow \cite{kam2015rviz}. Uber and GM Cruise launched an open-source visualization toolkit called Autonomous Visualization System (AVS), supporting building applications from autonomous and robotics data 
	\cite{AutonomousVisualizationSystem_2019}.  Apollo Simulation platform provides similar functions to enable 3D visualization \cite{apollo}.
	%Some footage of Rviz, AVS and Apollo Simulation is shown in Figure \ref{fig:vis}.		

\subsection{Benchmarks and Datasets} % copy here				
	
\begin{table*}[htp] 
	\caption{Open motion datasets for planning research, inspired by \cite{interactiondataset} }
\begin{center}
	\begin{tabular}{ | c | c | c | c | c | c | c |} %% TODO: open-source
		\hline
		%			Simulator & Traffic Flow  & Interactive  & Perception & Vehicle \\ 
		%					 &  & Road User  & input & Dynamics \\ \hline \hline 	& \makecell[l]{position error \\less than 10 cm } 	
		\makecell[c]{Dataset} & \makecell[c]{Viewpoint} & \makecell[c]{Country} & \makecell[c]{RT} & \makecell[c]{RoW} & \makecell[c]{Interactive Scenarios} & \makecell[c]{Highlights}   \\ \hline \hline
		
		%  & \makecell[c]{Accuracy} 		& \makecell[l]{mostly acceptable;\\ errors exist \cite{coifman2017critical}}
		NGSIM \cite{alexiadis}  & \makecell[l]{bird's-eye-view\\from a building} & USA & ST & E & \makecell[l]{ramp merging\\(double) lane change} & \makecell[l]{popularity}
		\\ \hline 
		
		highD \cite{highDdataset}  & \makecell[l]{bird's-eye-view\\from a drone} & DEU & ST & E &  \makecell[l]{lane change}
		  & \makecell[l]{large-scale,\\precision \& variety} \\ \hline 
		CITR/DUT \cite{DongfangYang}  & \makecell[l]{bird's-eye-view\\from a drone} & USA/CHN & UST & I & \makecell[l]{roundabouts\\unsignalized intersections\\pedestrian crossing}  & \makecell[l]{interpersonal \& vehicle-crowd \\ interaction
		} \\ \hline 		
		Stanford \cite{robicquet2016learning}  & \makecell[l]{bird's-eye-view\\from a drone} & USA & UST & I &  \makecell[l]{unsignalized intersections,\\pedestrian crossings} & \makecell[l]{large-scale,\\ various traffic-agents} \\ \hline 
		PREVENTION \cite{prevention_dataset}  & onboard sensors & ESP & ST & E & \makecell[l]{ramp merging \\(double) lane change\\pedestrian crossing} & \makecell[l]{data redundancy\\lane markings}
		 \\ \hline 		
		Argoverse \cite{chang2019argoverse}  & onboard sensors
		 & USA & UST & I &  \makecell[l]{unsignalized intersections\\pedestrian crossings} & \makecell[l]{rich map information}
		  \\ \hline 			
		INTERACTION \cite{interactiondataset}  & \makecell[l]{bird's-eye-view\\from a drone}  & \makecell{USA/CHN/\\DEU/BGR}
		 & UST & I & \makecell[l]{roundabouts, ramp merging,\\double lane change\\unsignalized intersections}  & \makecell[l]{diversity, complexity, \\criticality, semantic map} \\ \hline 
		PKU \cite{Zhao2017} & onboard sensors & CHN & ST & E & \makecell[l]{lane change} & \makecell[l]{high quality}\\ \hline 
		ApolloScape \cite{wang2019apolloscape}  & onboard sensors & CHN  & ST & E & \makecell[l]{(double) lane change, merging,\\ intersections, pedestrian crossings} 
		 & \makecell[l]{large \& rich labeling, \\ various traffic-agents}\\ \hline 
		HDD \cite{ramanishka2018toward}  & onboard sensors & USA  & ST & E & \makecell[l]{(double) lane change, merging,\\ intersections, pedestrian crossings} 
& \makecell[l]{ a novel annotation method,
	\\ various traffic-agents}\\ \hline 		
							
	\end{tabular}
		    \begin{tablenotes}
	\small
	\item The abbreviations are: RT (Road Type), RoW (Right of Way), ST (Structured), UST (Unstructured), E (explicit), I (implicit).   
\end{tablenotes}
	\label{tab:datasets}
\end{center}
\end{table*}
	
	Many benchmark suites or datasets for perception in automated driving exist (e.g. KITTI \cite{geiger2012we}). However, there are less benchmarks targeted for planning. To test the speedup techniques for routing, many researchers take advantage of available continent-sized benchmark instances, like the road network of Western Europe from PTV AG or the TIGER for USA road network \cite{Bast.20.04.2015}. Considering motion planning, one big challenge is that road scenarios lack reproducibility. Attempting to address this issue, CommonRoad was proposed by Althoff et.al. It is a benchmark collection for motion planning of road vehicles, comprising many advantages, such as reproducibility, composability, openness etc.\cite{Althoff.11.06.201714.06.2017}. Automated Driving Toolbox of MathWorks \cite{matlab} and Apollo Simulation platform \cite{apollo} has similar functions, which are able to generate testing scenarios and grade planning algorithms with respect to metrics. Some benchmarks are integrated in simulators. For example, The CARLA Autonomous Driving Challenge provides realistic traffic situations and ranks participated algorithms according to performance metrics \cite{dosovitskiy2017carla}. 
		
	Human interactions make the planning more difficult and should be understood by the algorithm. Thus, motion datasets are essential to understand the behavior and motion of other road users, which facilitate behavior-aware planning and enable validation of algorithms considering human interactions. The Next Generation Simulation (NGSIM) \cite{alexiadis} is one of the largest open datasets of naturalistic driving and is commonly used in many behavior related research, e.g. \cite{altche2017lstm,deo2018convolutional,Hou2019}. In recent years, a clear trend is the literacy in automated driving research. Numerous companies or institutes make their motion datasets public, as summarized in Table \ref{tab:datasets}. 
		
	\subsection{Middleware} % copy here
	A middleware is different from an operating system in the traditional sense, but offers services designed for a heterogeneous computer cluster. The open-source ROS \cite{quigley2009ros} is a robotics middleware, which includes many planning packages (see Section \ref{ssec:planningStacks}). Another middleware from the robotics domain is OROCOS \cite{OROCOS}. In automotive industry, the middleware ADTF (automotive data and time-triggered framework) is widely used \cite{messner2015eb}. Communication middleware like DDS and ZeroMQ used in distributed information systems are also useful for autonomous driving \cite{Watzenig.2017}.

	\section{Coupling between Tools} \label{Coupling of Tools} % Note update here
% Table generated by Excel2LaTeX from sheet 'Sheet1'
\begin{table*}[htp]

	\caption{Coupling between open-source tools for planning}
	\begin{center}
	\begin{tabular}{|c|P{0.7cm}|P{0.7cm}|P{0.7cm}|P{0.7cm}|P{0.7cm}|P{0.7cm}|P{0.7cm}|P{0.7cm}|P{0.7cm}|P{0.7cm}|P{0.7cm}|P{0.7cm}|P{0.7cm}|c|}
		\hline
		Tool & \multicolumn{6}{c|}{Simulator}                        & \multicolumn{1}{c|}{V2X} & \multicolumn{3}{c|}{Map} & \multicolumn{2}{c|}{Stack} & BM & MW   \\\hline
		\hline
		-  & Car   & LGS   & Gaz   & USA   & Air   & Sum  & Omn   & OSM   & Ope   & lan   & Aut   & Apo   & Com   & ROS \\
		\hline
		Carla & -     & o     & o     & o     & o     & o  &  o   &  o     & $\surd$  & o     & $\surd$  & $\surd$  & o     & $\surd$ \\
		\hline
		LGSVL & o     & -     & o     & o     & o     & o     & o     & o     & $\surd$  & $\surd$  & $\surd$  & $\surd$  & o     & $\surd$ \\
		\hline
		Gazebo & o     & o     & -     &\cite{shimizu2015realistic} & o     &\cite{garzon2018hybrid} & o     & $\surd$  & o     & o     & $\surd$  & o     & o     & $\surd$ \\
		\hline
		USARSim & o     & o     &\cite{shimizu2015realistic} & -     & o     &\cite{sumoUSARSim} &   o    & o     & o     & o     & o     & o     & o     & $\surd$ \\
		\hline
		AirSim & o     & o     & o     & o     & -     & o     & o     & o     & o     & o     & o     & o     & o     & $\surd$ \\
		\hline

		Sumo  & o  & o     &\cite{garzon2018hybrid} &    \cite{sumoUSARSim}   &  o     &  -   & \cite{Sommer}     & $\surd$  & $\surd$  &\cite{klischat2019coupling} & o     & o     &\cite{klischat2019coupling} &\cite{garzon2018hybrid} \\
		\hline
		Omnet++ & o     & o     & o     & o     & o     &\cite{Sommer} & -     & o     & o     & o     & o     & o     & o     & o \\
		\hline
		OSM   & o     & o     & $\surd$  & o     & o     & $\surd$  & o     & -     & o     & $\surd$  & $\surd$  & o     & $\surd$  & $\surd$ \\
		\hline
		OpenDrive & $\surd$  & $\surd$  & o     & o     & o     & $\surd$  & o     & o     & -     &\cite{althoff2018automatic} & $\surd$  & $\surd$  & $\surd$  & $\surd$ \\
		\hline
		lanelet1/2 & o     & $\surd$  & o     & o     & o     &\cite{klischat2019coupling} & o     & $\surd$  & \cite{althoff2018automatic} & -     & $\surd$  & o     & o     & $\surd$ \\
		\hline
		Autoware & $\surd$  & $\surd$  & $\surd$  & o     & o     & o     & o     & $\surd$  & $\surd$  & $\surd$  & -     & o     & o     & $\surd$ \\
		\hline
		Apollo & $\surd$  & $\surd$  & o     & o     & o     & o     & o     & o     & $\surd$  & o     & o     & -     & o     & $\surd$ \\
		\hline
		\makecell{Common\\-Road} & o     & o     & o     & o     & o     &\cite{klischat2019coupling} & o     & $\surd$  & $\surd$  & o     & o     & o     & -     & o \\
		\hline
	\end{tabular}%

	\begin{tablenotes}
	\small
	\item The second row refers to the same name as the first column, e.g. Car is the abbreviation of Carla. BM and MW denote Benchmark and Middleware respectively. The used symbols should be interpreted as: ($\surd$) interface available, (-) the same tool, and (o) not related or not found. The references of some independent interfaces are also given in the table.
	\end{tablenotes}	

	\label{tab:tools}%
	\end{center}
\end{table*}%
			
	In the previous Section, different types of tools assisting planning are reviewed. However, there is no versatile software package that is able to meet all development, validation, and deployment requirements for planning with respect to different inputs and application scenarios. Thus, a simulation framework incorporating tools from different domains is not only necessary but also avoids reinventing the wheel. Some commercialized co-simulation platforms, e.g. Model.CONNECT, can realize interdisciplinary simulation and facilitate function development for automated driving \cite{solmaz2019novel}. Another way to achieve co-simulation for scientific research is utilizing open interfaces. The look-up Table \ref{tab:tools} shows the current status of coupling between aforementioned open-source tools for planning.

	\section{CONCLUSION} \label{CONCLUSION}	
	
	Planning algorithms have been heavily surveyed in different literature (see Section \ref{DEFINITIONS}), but no publication is targeted at comprehensively reviewing tools supporting planning. On the other hand, the significance of tools (software infrastructure) have already been proven since DARPA urban challenge in 2007 \cite{mcnaughton2008software}. To fill this gap, this paper has defined planning and its required tools in the field of automated driving, and has clearly displayed the frontiers of various tools supporting planning and the bridges between them. The aim of the paper is to help researchers to make full use of open-source resources and reduce effort of setting up a software platform that suites their needs. For example, a reader attempts to develop a novel motion planner. It is a good option to choose open-source Autoware as software stack along with ROS middleware, as Autoware can be further transferred to a real vehicle. During the development, he or she can use Carla as a simulator, to get its benefits of graphic rendering and sensor simulation. To make the simulation more realistic, he or she might adopt commercial software CarMaker for sophisticated vehicle dynamics and open-source SUMO for large-scale traffic flow simulation. OpenDRIVE map can be used as a source and converted into the map format of Autoware, Carla and SUMO. Finally, CommonRoad can be used to evaluate the developed algorithm and benchmark it against other approaches. Considering coupling between tools, either co-simulation software or open interfaces in Table \ref{Coupling of Tools} can help. Hence tedious work on developing tools as well as interfaces can be reduced and more focus can be put on the algorithm development. 
	 
	Furthermore, despite the fact that many advances of supporting tools have emerged in recent years, we believe that the following gaps need to be filled in the future:
	\begin{itemize}
		\item Benchmarks for routing in a dynamic transportation network and for behavior planning are not available. Therefore, a quantitative comparison remains important further work.
		\item All benchmarks attach importance to "performance" of an AV, rather than "human emotions". It is still a question how to evaluate a planing algorithm with human judgment, not only from the vehicle occupants but also from other road users.
		\item Many open motion datasets have been made public in the past few years. However, they have various data formats, leading to the difficulty of general use. A standard format for motion datasets is expected to be proposed. %Also, only a little work puts the realistic datasets in their alf, e.g. highD dataset used for design and testing of ADAS\cite{Schiegg.22.09.201925.09.2019}. More effort should be put into Dataset-in-the-loop for validating algorithms with realistic traffic data.	
%		\item Cooperation among different tools is necessary in order to take advantages of %tools from different domains. However, plenty of interfaces are missing in Table %\ref{tab:tools}, especially between communication (V2X) and other domains.
	\end{itemize}

	\addtolength{\textheight}{-0cm}   % This command serves to balance the column lengths
	% on the last page of the document manually. It shortens
	% the textheight of the last page by a suitable amount.
	% This command does not take effect until the next page
	% so it should come on the page before the last. Make
	% sure that you do not shorten the textheight too much.
	
	%%%%%%%%%%%%%%%%%%%%%%%%%%%%%%%%%%%%%%%%%%%%%%%%%%%%%%%%%%%%%%%%%%%%%%%%%%%%%%%%

	%%%%%%%%%%%%%%%%%%%%%%%%%%%%%%%%%%%%%%%%%%%%%%%%%%%%%%%%%%%%%%%%%%%%%%%%%%%%%%%%

	%%%%%%%%%%%%%%%%%%%%%%%%%%%%%%%%%%%%%%%%%%%%%%%%%%%%%%%%%%%%%%%%%%%%%%%%%%%%%%%%
%	\section*{APPENDIX}
	
%	Appendixes should appear before the acknowledgment.
	
	\section*{ACKNOWLEDGMENT}

	The authors would like to thank all national funding authorities 	and the ECSEL Joint Undertaking, which funded the NewControl project under the grant agreement No. 826653-2.\par
	The publication was written at VIRTUAL VEHICLE Research GmbH in Graz and partially funded by the COMET K2 - Competence Centers for Excellent Technologies Programme of the Federal Ministry for Transport, Innovation and Technology (bmvit), the Federal Ministry for Digital and Economic Affairs (bmdw), the Austrian Research Promotion Agency (FFG), the Province of Styria and the Styrian Business Promotion Agency (SFG).\par

	%%%%%%%%%%%%%%%%%%%%%%%%%%%%%%%%%%%%%%%%%%%%%%%%%%%%%%%%%%%%%%%%%%%%%%%%%%%%%%%%

	\bibliographystyle{ieeetr}
	\bibliography{myRef}

\end{document}